\documentclass[11pt,a4paper]{article}
\usepackage[hyperref]{acl2021}
\usepackage{times}
\usepackage{latexsym}
\usepackage{graphicx}
\usepackage{amsmath}
\usepackage{amsfonts}
\usepackage{amssymb}
\usepackage{multirow}
\usepackage{makecell}
\usepackage{booktabs}
\usepackage{verbatim}

\usepackage{microtype}

\aclfinalcopy 

\title{STN4DST: A Scalable Dialogue State Tracking based on Slot Tagging Navigation}

\author{Puhai Yang, Heyan Huang, Xian-Ling Mao \\
	Beijing Institute of Technology \\
	\texttt{\{phyang, hhy63, maoxl\}@bit.edu.cn}}

\date{}

\begin{document}

\maketitle

\begin{abstract}
Scalability for handling unknown slot values is a important problem in dialogue state tracking (DST). As far as we know, previous scalable DST approaches generally rely on either the candidate generation from slot tagging output or the span extraction in dialogue context. However, the candidate generation based DST often suffers from error propagation due to its pipelined two-stage process; meanwhile span extraction based DST has the risk of generating invalid spans in the lack of semantic constraints between start and end position pointers. To tackle the above drawbacks, in this paper, we propose a novel scalable dialogue state tracking method based on slot tagging navigation, which implements an end-to-end single-step pointer to locate and extract slot value quickly and accurately by the joint learning of slot tagging and slot value position prediction in the dialogue context, especially for unknown slot values. Extensive experiments over several benchmark datasets show that the proposed model performs better than state-of-the-art baselines greatly\footnote{This work has been submitted to the IEEE for possible publication. Copyright may be transferred without notice, after which this version may no longer be accessible.}.
\end{abstract}

\section{Introduction}
Dialogue state tracking (DST) plays a key role in task-oriented dialogue systems to track the user's intentional state and convert it into a set of slot-value pairs, i.e., dialogue state. As the basis for selecting the next system action, the accurate prediction of dialogue state is critical. Traditionally, DST approaches typically assumes that all candidate slot-value pairs are available in advance, and then a slot-value pair is selected as the predicted one by scoring all slot-value pairs or performing classification over the set of all slot values \cite{mrkvsic2017neural,ren2018towards,zhong2018global,lee2019sumbt,zhang2019find,shan2020contextual}. Since in real scenario it is often impossible to know all the candidate slot values in advance  \cite{rastogi2017scalable,xu2018end}, these traditional methods, also known as fixed ontology-based DST, are often ineffective due to their inability to scale to unknown slot values.

Recently, open vocabulary-based DST attracts increasing attention, which is independent of fixed candidate slot value ontology and can scale to unknown slot values. In previous open vocabulary-based DST, two types of methods are primarily concerned: candidate generation based DST and span extraction based DST. Candidate generation based DST depends on language understanding or N-gram to generate the list of candidate slot values, and then scores these candidate slot values to select the predicted slot values \cite{rastogi2017scalable,goel2018flexible,rastogi2018multi}. Span extraction based DST is now more widely used because this kind of DST method can efficiently extract slot values by simply generating the slot value's start and end positions in dialogue context \cite{xu2018end,chao2019bert,gao2019dialog,zhang2019find,heck2020trippy}.

However, both kinds of methods above have great defects because of their own characters. For candidate generation based DST, many previous works have pointed out that it is prone to unavoidable error propagation and low efficiency \cite{rastogi2017scalable,xu2018end,chao2019bert} due to its pipelined two-stage process, i.e., extract the candidate and then score them. For span extraction based DST, there is a lack of semantic constraint between the start and end pointers of the slot values generated by the span extraction. Especially for unknown slot values containing multiple words, span extraction results in serious invalid spans because it cannot effectively utilize the context semantic information that constrains start-end pointer pairs.

To tackle the above drawbacks, in this paper, we propose a novel scalable dialogue state tracking method based on slot tagging navigation (STN4DST), which implements an end-to-end single-step pointer to locate and extract slot value quickly and accurately by the joint learning of slot tagging and slot value position prediction, especially for unknown slot values. Specifically, all candidate slot values in a dialogue are first located by slot tagging. Then, for each slot, a distribution is generated on the dialogue to obtain the slot value start position pointer. Finally, the slot value is extracted directly from the dialogue using the start position pointer and slot tagging output. Note that, previous dialogue state and special slot values (e.g., $dontcare$) are added to the input of STN4DST to achieve automatic dialogue state update and implicit slot value prediction.

The contributions of our work are as follows:
\begin{itemize}
	\item We propose a novel scalable dialogue state tracking method based on slot tagging navigation (STN4DST), which can locate and extract slot value quickly and accurately, especially for unknown slot values.
	\item Our model achieves state-of-the-art performance on several benchmark datasets in an open vocabulary-based DST setting.
	\item Our model shows the potential for handling unknown slot values containing multiple words, which is a limitation in previous DST approaches.
\end{itemize}

\section{Related Work}
Previous open vocabulary-based dialogue state tracking (DST) can be divided into two categories according to the different ways of predicting slot value: candidate generation based DST  \cite{rastogi2017scalable,goel2018flexible,rastogi2018multi} and span extraction based DST \cite{xu2018end,chao2019bert,gao2019dialog,zhang2019find,heck2020trippy}. The former usually requires a candidate list of slot values, which can be either an N-gram list on the dialogue context or the slot tagging output of language understanding. The latter is now more widely used to extract slot values from dialogue context by directly generating start and end position pointers. In detail, the both kinds of DST methods above can be introduced as follows:

\paragraph{Candidate Generation based DST:} \citeauthor{rastogi2017scalable} (\citeyear{rastogi2017scalable}) extract utterance related, slot related and candidate related features and feed them into slot specific candidate scorer to update the score of each candidate in the candidate set, the list of candidates they use comes from ground truth slot tagging label. \citeauthor{goel2018flexible} (\citeyear{goel2018flexible}) create two kinds of candidate sets: N-gram candidate set and spoken language understanding (SLU) candidate set. They produces a binary classification decision for each combination of a candidate value and a slot type to determine whether the candidate is the update value of the slot type or not. \citeauthor{rastogi2018multi} (\citeyear{rastogi2018multi}) use slot values obtained from slot tagging to update the set of candidates for each slot in each turn, and depend on parameter-shared slot scorer to score all candidate in each slot.

\paragraph{Span Extraction based DST:} \citeauthor{xu2018end} (\citeyear{xu2018end}) first introduce pointer network \cite{vinyals2015pointer} into DST. They use a bidirectional recurrent neural network to encode all dialogue history to obtain a context representation, which is then fed into a gate to determine whether to extract slot values. And the slot value is extracted by generating the start and end pointers in the dialogue history. \citeauthor{gao2019dialog} (\citeyear{gao2019dialog}) formulate DST as a reading comprehension task to answer the question $``what\ is\ the\ state\ of\ the\ current\ dialog?"$ after reading dialogue context. And they use a simple attention-based neural network to point to the start and end positions of slot values within the dialogue. \citeauthor{chao2019bert} (\citeyear{chao2019bert}) use BERT \cite{devlin2018bert} as dialogue context encoder to obtain semantic context, so the span prediction module can more effectively predict the starting and ending positions of slot values. \citeauthor{zhang2019find} (\citeyear{zhang2019find}) target the issue of illformatted strings that generative models suffer from and take a hybrid approach for picklist-based slots and span-based slots. \citeauthor{heck2020trippy} (\citeyear{heck2020trippy}) use a triple copy strategy to fill slots in the dialogue state. They set up three copy mechanisms for slot values: span prediction on dialogue context, copy from system inform memory and coreference resolve on previous dialogue state.

\section{The Proposed Model}
\begin{figure*}[h]
	\centering
	\includegraphics[width=\linewidth]{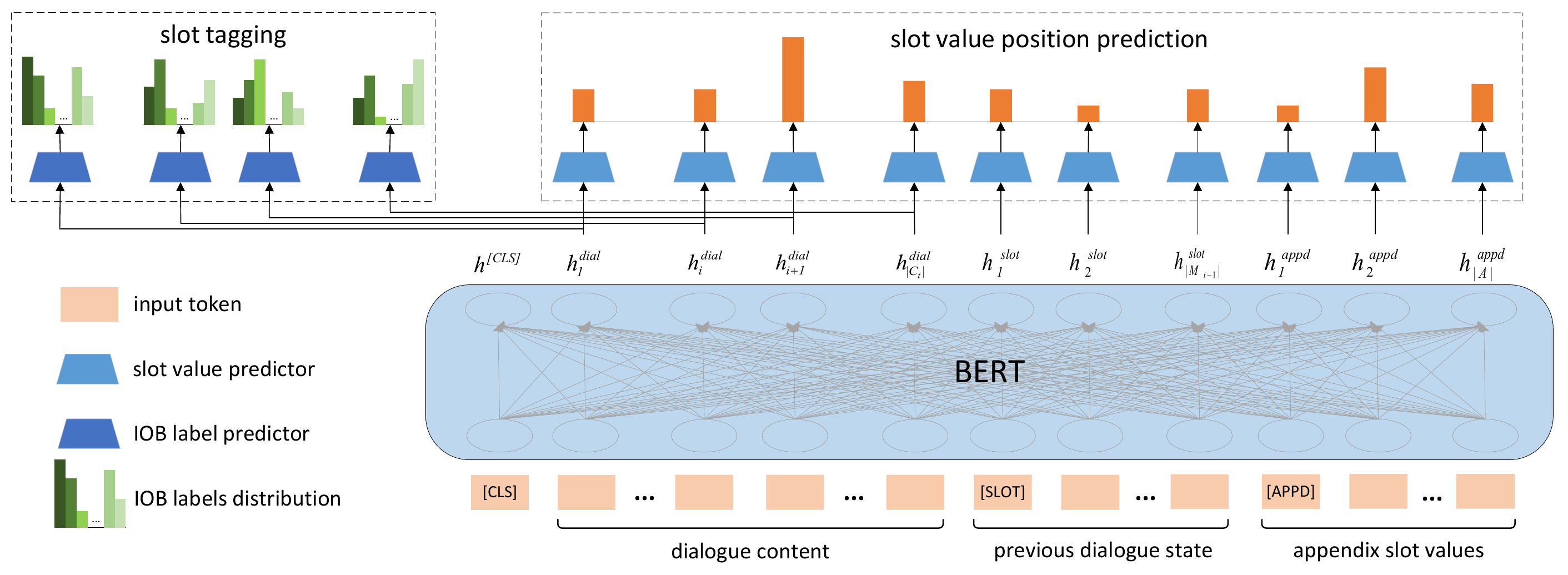} 
	\caption{Architecture of the proposed STN4DST. For each turn, STN4DST takes the dialogue content, the previous dialogue state and appendix slot values as input, and outputs the IOB labels of the dialogue content and the position of each slot value in the input. Parameters are shared in BERT and IOB label predictor for all slots, while the slot specific prediction layer is used for the slot value predictor.}
	\label{framework}
\end{figure*}

The architecture of the proposed model\footnote{Our code will be released upon publication of this work.} is shown in Figure \ref{framework}. Our model consists of three parts: dialogue encoder, slot tagging module and slot value position prediction module.

In our model, the dialogue encoder takes the dialogue content, the previous dialogue state and appendix slot values as input to get the contextual representation. Guided by the output of the encoder, the slot tagging is used to locate all candidate slot values in the dialogue content; meanwhile, in the slot value position prediction, a distribution on the output is generated for each slot to navigate to the position of the predicted slot value.

\subsection{Dialogue Encoder}
We use pretrained BERT \cite{devlin2018bert}, which has been proved to work well in natural language representation, as encoder in our model. The input of the encoder consists of three parts: the dialogue content, the previous dialogue state and appendix slot values. Now, let's introduce the format definitions of the three different parts.

\paragraph{Dialogue Content:} Define $D=\{(Y_1, U_1),...,(Y_T,U_T)\}$ as the dialogue sequence of length $T$, where $Y_t$ is the system utterance at turn $t$, and $U_t$ is the user utterance following the system utterance. The dialogue content at turn $t$ in our model is $C_t=H_t^{l}\oplus{[SEP]}\oplus{Y_t}\oplus{;}\oplus{U_t}\oplus{[SEP]}$ where $H_t^{l}=\{(Y_{t-l}, U_{t-l}),...,(Y_{t-1},U_{t-1})\}$ is the dialogue history up to turn $t$ of length $l$, character ``$;$" is a special token used to mark the boundary between $Y_t$ and $U_t$, and $[SEP]$ is a special token for separating different sentences in BERT. Hyperparameter $l$ is usually set to 0 or 1, and a proper length of dialogue history is necessary in some cases for slot value references that may occur across multiple turns of dialogues.

\paragraph{Previous Dialogue State:} Let $M_t=\{(S^n,V_t^n)|1\le{n}\le{N}\}$ be the dialogue state at turn $t$, $N$ is the number of slot in the predefined state ontology, $S^n$ is the $n$-th slot and $V_t^n$ is the value of the slot at turn $t$. To implement the dialogue state update mechanism, we define the dialogue state as a fixed-size memory, referring to the setting in SOM-DST \cite{kim-etal-2020-efficient}. Thus, the previous dialogue state in the encoder input at turn $t$ is denoted as $M_{t-1}=M_{t-1}^1\oplus{}...\oplus{}M_{t-1}^N$, in which $M_{t-1}^n=[SLOT]\oplus{}S^n\oplus{}-\oplus{}V_{t-1}^n$ is the representation of slot-value pair. ``$-$" is a special token used to mark the boundary between a slot and a value, $[SLOT]$ is a special token used to aggregate the information of the slot-value pair into a single vector. In addition, if the slot value of a slot is not mentioned before turn $t$, we use a uniform token $[NULL]$ to represent the slot value.

\paragraph{Appendix Slot Values:} Some special slot values cannot be extracted directly from the dialogue content, for example, $dontcare$ indicates that the user does not have specific requirements for a slot, which often needs to be inferred from the dialogue content. In previous works \cite{xu2018end,chao2019bert,gao2019dialog,heck2020trippy,kim-etal-2020-efficient}, these special slot values are usually solved by the gate mechanism of slot value type classification, which ignores the semantic information of these special values. In this paper, we add these special values as appendix slot values to the input of the encoder to make more efficient use of the semantic information of the values. Specifically, let $\{A^1,...,A^k\}$ be the set of all $k$ special values, the appendix slot values in the encoder input are concatenated into $A=[APPD]\oplus{}A^1...[APPD]\oplus{}A^k$, in which $[APPD]$ is a special token used to aggregate the information of the special value into a single vector, same as $[SLOT]$. 

The input tokens of the encoder are the concatenation of the three mentioned above:
$$
	R=[CLS]\oplus{}C_t\oplus{}M_{t-1}\oplus{}A
$$
where $[CLS]$ is the special token placed at the beginning of the input in BERT.

The input of BERT is the concatenation of the WordPiece embedding \cite{wu2016google} for input tokens, segment embedding, and positional embedding. For the segment embedding, we use 0 for the tokens that belong to $H_t^{l}$ and 1 for all other tokens, the positional embedding follow the settings in BERT.

The embedded input sequence is then passed to BERT’s bidirectional Transformer encoder, and the final hidden states are denoted as:
$$
\begin{aligned}
	&[h^{[CLS]},h_1^{dial},...,h_{|C_t|}^{dial},h_1^{slot},...,h_{|M_{t-1}|}^{slot},\\
	&h_1^{appd},...,h_{|A|}^{appd}]
\end{aligned}
$$
where $h^{[CLS]}$ is a representation of the entire input, $H^{dial}=[h_1^{dial},...,h_{|C_t|}^{dial}]$, $H^{slot}=[h_1^{slot},...,h_{|M_{t-1}|}^{slot}]$ and $H^{appd}=[h_1^{appd},...,h_{|A|}^{appd}]$ are contextual representations for the dialogue content, the previous dialogue state and appendix slot values, respectively. $H^{dial}$ is used for slot tagging to locate all candidate slot values in dialogue content, and each slot in the dialogue state is determined by a distribution on the concatenation of $H^{dial}$, $H^{slot}$ and $H^{appd}$.

\subsection{Slot Tagging for Dialogue Content}
Slot tagging is introduced in our model as multi-task learning, and its ability to solve unknown slot values has been demonstrated in previous works. Specifically, in previous studies \cite{goel2018flexible,rastogi2018multi}, slot tagging is usually used to extract candidate slot values from a dialogue and add them to candidate set, and then semantic matching is adopted to realize slot value selection. However, the above approach not only cause error propagation, but also lost the context semantic information of the slot value in the dialogue. In this paper, we propose a new multi-task learning strategy, called slot tagging navigation, to joint learning slot tagging and slot value position prediction.

Let $S=\{S^1,...,S^N\}$ be the set of all slots in the predefined state ontology, a set of $2|S|+1$ labels ($-B$ label and $-I$ label for each slot and a single $O$ label) is defined for IOB tagging scheme \cite{sang2000introduction}. For each token representation $h_i^{dial}$ in dialog content representation $H^{dial}$, it is linearly projected through a common layer, IOB label predictor, whose output values correspond to all IOB tagging labels. Softmax is then applied to the output values to produce a probability distribution across all IOB labels, by which the IOB label of $i$-th token in dialogue content can be determined:
$$
	P_i^{iob}(h_i^{dial})=softmax(W^{iob}\cdot{}h_i^{dial}+b^{iob}) \in{}\mathbb{R}^{2|S|+1}
$$
where $W^{iob}$ and $b^{iob}$ are trainable parameters trained from scratch in our model.

Different from previous works where $P_i^{iob}$ is used only to extract candidate slot values, in slot tagging navigation, we assume that each token in the dialogue content is the starting position of a candidate slot value, and $P_i^{iob}$ is used to determine the specific range of the candidate slot value. In addition, the slot tagging task also performs the function of aggregating the information of the candidate slot value into the token with the $-B$ label. So, we can use each token representation $h_i^{dial}$ to represent a candidate slot value, thereby avoiding error propagation and obtaining context semantic information. 

\subsection{Single-step Slot Value Position Prediction}
For open vocabulary-based dialogue state tracking, the span extraction has attracted great attention because of its ability to extract unknown slot values from the dialogue. However, this kind of methods require two-step prediction of slot value starting and ending positions and lacks semantic constraints between the two steps, which makes this method inefficient and unwarranted. In this paper, benefiting from slot tagging navigation's assumption that each token in the dialogue content is the starting position of a candidate slot value, we are able to predict the starting position of slot value by simply generating a distribution on the dialogue content, thus achieving single-step slot value position prediction.

In addition to the dialogue content, the previous dialogue state and appendix slot values are added to the input of slot value predictor to learn the dialogue state update mechanism and track implicit slot values. The input $V$ of our slot value predictor is the concatenation of $H^{dial}$, $H^{slot}$ and $H^{appd}$:
$$
\begin{aligned}
	V=&[h_1^{dial},...,h_{|C_t|}^{dial},h_1^{slot},...,h_{|M_{t-1}|}^{slot},\\
	&h_1^{appd},...,h_{|A|}^{appd}]
\end{aligned}
$$
For each slot $S^n$ that is to be predicted, a slot specific prediction layer takes each token representation $v_j$ in $V$ as input and projects them as follows:
$$
	\alpha_j^n=W_n^{value}\cdot{}v_j+b_n^{value} \in{}\mathbb{R}^{1}
$$
$$
	P_n^{value}=softmax(\alpha^n)
$$
$$
	position_n^{value}=argmax(P_n^{value})
$$
where $W_n^{value}$ and $b_n^{value}$ are trainable parameters for each slot $S^n$.

$position_n^{value}$ is a pointer to the input $V$, which may hit any position in $V$. Corresponding to the input of the encoder, according to the $position_n^{value}$ hit position, slot value extraction can be divided into three cases as follows:

\paragraph{Hit on Dialogue Content:} This situation indicates that the slot value needs to be extracted from the dialogue. According to the slot tagging output, if the $position_n^{value}$ hit position is marked with $-B$, the slot value hit successfully, and it is extracted using the IOB tagging rule.

\paragraph{Hit on Previous Dialogue State:} The slot retains the previous value or references the value of another slot. A successful hit occurs when the $position_n^{value}$ points to any $[SLOT]$ in the input, if the slot hits itself, it retains its previous value; if it hits other slots, coreference resolution \cite{heck2020trippy} occurs.

\paragraph{Hit on Appendix Slot Values:} The slot value is a special value that cannot be extracted directly from the dialogue but needs to be inferred. When the $position_n^{value}$ points to any $[APPD]$ in the input, the special value following it is taken as the predicted slot value to form a successful hit.

Note that when the $position_n^{value}$ does not hit any valid token (with $-B$ label, or $[SLOT]$, or $[APPD]$), the slot value will not be updated. In practice, the $position_n^{value}$ can always hit successfully. Since there are very few slot values to be updated each turn, the slot value pointer tends to hit its own $[SLOT]$ token to retain the previous value,

\subsection{Objective Function}
In the training, slot tagging and slot value position prediction are jointly optimized by fine-tuning on BERT.

\paragraph{Slot Tagging:} The slot tagging loss for all tokens in dialogue content is the average of the negative log-likelihood, as follows:
$$
	Loss^{iob}=-\frac{1}{I}\sum_{i=1}^I(Y_i^{iob})^\mathsf{T}log(P_i^{iob})
$$
where $Y_i^{iob}$ is the one-hot vector for the ground truth IOB label for the $i$-th token in dialogue content.

\paragraph{Slot Value Position Prediction:} The objective function for slot value position prediction is also the average of the negative log-likelihood:
$$
	Loss^{value}=-\frac{1}{N}\sum_{n=1}^N(Y_n^{value})^\mathsf{T}log(P_n^{value})
$$
where $Y_n^{value}$ is the one-hot vector for the ground truth slot value hit position for the $n$-th slot.

Therefore, the final joint objective function to be minimized is the sum of the above two loss functions:
$$
	Loss^{joint}=Loss^{iob}+Loss^{value}
$$

\section{Experiment Settings}

\subsection{Datasets}

\begin{table}[!t]
	\centering
	\begin{tabular}{ccccc}
		\hline
		Dataset & \# Slots & \makecell[c]{\# Dialogues \\ (train, dev, test)} & USV\% \\
		\hline
		Sim-M & 5 & 384, 120, 264  & 33.3\% \\
		Sim-R & 9 & 1116, 349, 775  & 12.5\% \\
		WOZ2.0 & 3 & 600, 200, 400  & 3.7\% \\
		MWOZ2.2 & 11 & 8438, 1000, 999  & 22.5\% \\
		\hline
	\end{tabular}
	\caption{Data statistics of Sim-M, Sim-R, WOZ2.0 and MWOZ2.2. The last column presents the percentage of unique slot values in the test set that were not observed in the training data, i.e., unknown slot values.}
	\label{datasets}
\end{table}

We evaluate our model on four benchmark datasets: \textbf{Sim-M} \cite{shah2018building}, \textbf{Sim-R} \cite{shah2018building}, \textbf{WOZ2.0} \cite{wen2016network} and \textbf{MWOZ2.2} \cite{zang2020multiwoz}, the statistics of these datasets are shown in Table \ref{datasets}.

Sim-M and Sim-R are multi-turn dialogue datasets in the movie and restaurant domains. Since the percentage of unknown slot values (out-of-vocabulary values) in Sim-M's $movie$ slot and Sim-R's $restaurant\_name$ slot reaches 100\% and 39\% respectively, these two datasets present a challenge for dialogue state tracking to scale to unknown slot values. Span annotations for all slot values (except special value $dontcare$) in the system utterance and user utterance are provided in these two datasets so that full token-level IOB labels can be obtained in the training of slot tagging.

WOZ2.0 provides automatic speech recognition (ASR) hypotheses of user utterances that can be used to assess the robustness of dialogue state tracking against ASR errors, the dataset contains three slots for the restaurant domain: $food$, $area$, and $price\_range$. As in previous works, we use manuscript user utterance for training and top ASR hypothesis for testing. The WOZ2.0 dataset does not provide full token-level IOB labels, but only slot filling information for sentence-level language understanding (LU). Since such LU annotation is more common and accessible in practice, we extract the incomplete IOB labels directly from the LU annotation to verify the performance of our model in the case of low IOB label resources.

MWOZ2.2 is the latest revision of MultiWOZ \cite{budzianowski2018multiwoz}, a well-known task-oriented dialogue dataset containing over 10,000 annotated dialogues spanning 8 domains. MWOZ2.2 uses a predefined schema to divide slots into two categories: non-categorical and categorical, in which the slots with fewer than 50 different slot values in the training set were classified as categorical, and the others as non-categorical. We carried out automatic IOB tagging of all 11 non-categorical slots in the five frequently studied domains ($train$, $attraction$, $restaurant$, $hotel$, $taxi$) on the MWOZ2.2 dataset. It should be noted that the capacity of our model is limited by incomplete IOB labels.

Following previous works \cite{xu2018end,chao2019bert,heck2020trippy}, the \textbf{joint goal accuracy} on all test sets is calculated for comprehensive comparison with other approaches. Joint goal accuracy is the accuracy to check whether all the predicted slot values in a turn match the ground truth slot values exactly. In addition, we calculate \textbf{slot accuracy}, \textbf{slot tagging accuracy} and \textbf{slot value position prediction accuracy} to carry out detailed model analysis.

\subsection{Baselines}
We compare the performance of STN4DST with both fixed ontology-based DST and open vocabulary-based DST.

\paragraph{NBT-CNN}\cite{mrkvsic2017neural} uses CNN to obtain the semantic representation of dialogue context and slot-value pair candidate, and then selects the candidate to update the dialogue state by establishing a binary classifier.
\paragraph{SMD-DST} \cite{rastogi2017scalable} extracts feature from system and user utterances with a two layer stacked bi-directional GRU network and scores all candidates from the ground truth slot tagging output for each slot to select slot value.
\paragraph{LU-DST} \cite{rastogi2018multi} uses a bidirectional GRU to encode the user utterance and selects slot value for each slot by scoring the candidates from slot tagging output.
\paragraph{SpanPtr} \cite{xu2018end} encodes the whole dialogue history with a bidirectional LSTM and extracts slot value for each slot by generating the start and end positions in dialogue history.
\paragraph{TRADE} \cite{wu2019transferable}uses a slot gate to predict whether slot values need to be generated, and there is a PGN-based state generator in the model to generate slot values.
\paragraph{GLAD} \cite{zhong2018global} takes previous system action and current user utterance as the input of self-attentive RNNs, and then calculates the semantic similarity between the output of RNN and the items in the predefined ontology to select the predicted value.
\paragraph{StateNet} \cite{ren2018towards} applies LSTM to track the inner dialogue states among the dialogue turns. And for each slot, it outputs a corresponding probability distribution over the set of possible values at each of the dialogue turn.
\paragraph{SUMBT} \cite{lee2019sumbt} uses BERT to encode domain-slot pairs and dialogue contexts, and then relies on the distance measure to select slot values from the slot value ontology.
\paragraph{BERT-DST} \cite{chao2019bert} exploits BERT-base as the encoder for the dialogue context and extracts the value of the slots from the dialogue context as a span.
\paragraph{TripPy} \cite{heck2020trippy} encodes the whole dialogue context with BERT and sets up three copy mechanisms for slot values: span extraction on dialogue context, copy from system inform memory and coreference resolve on previous dialogue state.
\paragraph{SOM-DST}\cite{kim-etal-2020-efficient} uses an explicit memory that can be selectively overwritten to represent the dialogue state and divides the prediction for each slot into two steps: slot operation prediction and slot value generation.

\subsection{Training}
The pre-trained BERT \cite{vaswani2017attention} (BERT-Base, Uncased) which has 12 hidden layers of 768 units and 12 self-attention heads is used as encoder in our model. During training, we use the BertAdam optimizer \cite{kingma2014adam} and set the peak learning rate to 4e-5 and the warmup proportion to 0.1. We use a 10\% dropout \cite{srivastava2014dropout} rate and a batch size of 16. The max sequence length is set to 128 except for 150 on WOZ2.0 and 280 on MWOZ2.2. For dialogue content, we deploy word dropout \cite{bowman2016generating} by randomly replacing the tokens in it with special token $[UNK]$ with the probability of 0.1. In addition, on the Sim-M dataset, we use a value dropout \cite{xu2014targeted} with a probability of 0.4, which randomly replaces the slot value in the dialogue content with the special token $[UNK]$.

We train the model for 100 epochs, and the training is stopped early when the joint goal accuracy on dev set is not improved for 15 consecutive epochs. For more reliable results, all the reported results of the proposed model are averages over three runs with different seeds.

\section{Experimental Results}

\begin{table}[!t]
	\centering
	\begin{tabular}{ccc}
		\hline
		DST Model & Sim-M & Sim-R \\
		\hline
		SMD-DST  & 96.8\%$^\dagger$ & 94.4\%$^\dagger$ \\
		TripPy  & 83.5\%$^*$ & 90.0\%$^*$ \\
		\hline
		LU-DST  & 50.4\% & 87.1\% \\
		BERT-DST  & 80.1\% & 89.6\% \\
		\hline
		STN4DST & \textbf{85.4$\pm$1.4\%} & \textbf{95.4$\pm$0.2\%} \\
		\hline
	\end{tabular}
	\caption{Joint goal accuracy on Sim-M and Sim-R. $^\dagger$ indicates the corresponding model should be considered as a strong oracle because the candidates are ground truth slot tagging labels. $^*$ indicates the corresponding model should be considered as a weak oracle because the system action that contains the ground truth slot value is used as auxiliary feature.}
	\label{result_M2M}
\end{table}

\begin{table}[!t]
	\centering
	\begin{tabular}{cc}
		\hline
		DST Model & MWOZ2.2 \\
		\hline
		SpanPtr  & 38.4\% \\
		TRADE  & 62.4\% \\
		BERT-DST  & 65.4\% \\
		SOM-DST  & 69.6\% \\
		\hline
		STN4DST & \textbf{68.7$\pm$0.0\%} \\
		\hline
	\end{tabular}
	\caption{Joint goal accuracy on MWOZ2.2. The corresponding experimental results are obtained from the baseline model reproduced in this paper.}
	\label{result_MWOZ}
\end{table}

\begin{table}[!t]
	\centering
	\begin{tabular}{cc}
		\hline
		DST Model & WOZ2.0 \\
		\hline
		NBT-CNN  & 84.2\%  \\
		GLAD  & 88.1\%  \\
		StateNet  & 88.9\%  \\
		SUMBT  & 91.0\%  \\
		\hline
		TripPy  & 92.7\%$^*$  \\
		BERT-DST  & 87.7\%  \\
		\hline
		STN4DST & \textbf{89.4$\pm$0.2\%}  \\
		\hline
	\end{tabular}
	\caption{Joint goal accuracy on WOZ2.0. The top group of models require a fixed slot value ontology. $^*$ should be considered as a oracle because the system action that contains the ground truth slot value is used as auxiliary feature.}
	\label{result_WOZ2.0}
\end{table}

Tables \ref{result_M2M}, \ref{result_MWOZ}, and \ref{result_WOZ2.0} show the performance of our model compared to various baselines, it can be observed that:

In the case of high unknown slot value ratio, the performance of our model has a great absolute advantage over previous state-of-the-art baselines. For examples, on Sim-M dataset, our model achieves an absolute improvement of 5.3\% compared with BERT-DST. Even compared with the previous state-of-the-art model TripPy, which uses system action as an auxiliary feature, our model still exceeds it by 1.9\%. Over Sim-R dataset, we promote the joint goal accuracy to 95.4\%, an absolute improvement of 5.4\% compared with the best result published previously. On MWOZ2.2 dataset, our model still outperforms the span-based BERTDST model by 3.3\%, although the annotations are incomplete.

In the case of relatively few unknown slot values, our proposed model maintains comparable results. For examples, on WOZ2.0 dataset, our model achieves a joint goal accuracy of 89.4\%, and improves the performance by 1.7\% in the open vocabulary-based DST setting.


\subsection{Ablation Study}
Ablation experiments are conducted on Sim-M, shown in Table \ref{ablation}. Binary classifier (the value is $dontcare$ or not) is used in ``- appendix slot values" to classify the types of slot values, while triple classifier (the value is $dontcare$ or not mentioned, or extract from dialogue content) is used in the three different versions of the model below ``- appendix slot values". In addition, span extraction is used in ``- position prediction" and ``- slot tagging" for slot value extraction. 

It can be observed that appendix slot values and previous dialogue state all contribute to joint goal accuracy. For examples, removing appendix slot values reduces joint goal accuracy by 1.1\%, and the joint goal accuracy further decreases by 1.3\% without previous dialogue state.

\begin{table}[!t]
	\centering
	\begin{tabular}{lc}
		\hline
		Model & Sim-M \\
		\hline
		STN4DST & \textbf{85.4\%} \\
		{\quad- appendix slot values} & 84.3\% \\
		{\quad\quad- previous dialogue state} & 83.0\% \\
		{\quad\qquad- position prediction} & 79.1\% \\
		{\qquad\qquad- slot tagging} & 78.9\% \\
		\hline
	\end{tabular}
	\caption{Ablation experiments for our model in joint goal accuracy on Sim-M. We use a cascading strategy to gradually remove each part in the model. In particular, ``position prediction" indicates the single-step slot value position prediction module. The `` - " in each item represents the further removal of the item's module after removing the module in all items above. For example, ``- previous dialogue state" refers to removing both appendix slot values and previous dialogue state from the model input.}
	\label{ablation}
\end{table}

Moreover, slot tagging navigation makes the greatest contribution to our model. For example, after we further remove slot tagging navigation, the joint goal accuracy reduces by 4.1\%. In particular, removing only the single-step slot value position prediction in slot tagging navigation result in a 3.9\% drop in joint goal accuracy, suggesting that slot tagging navigation is a relatively better multi-task learning strategy joint with slot tagging in dialogue state tracking.

\subsection{Generalization Study}
\begin{figure}[t]
	\centering
	\includegraphics[width=\linewidth]{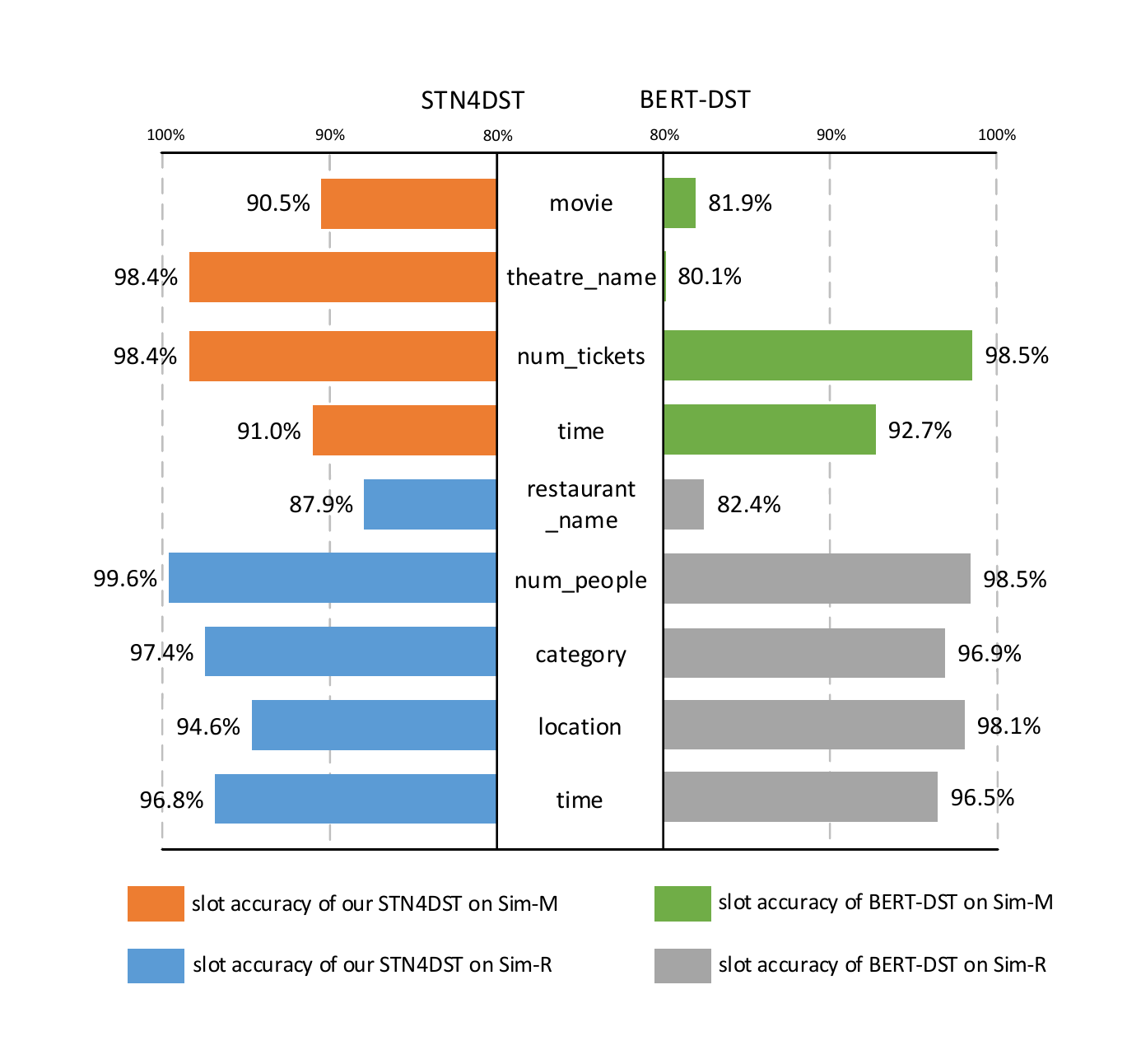}
	\caption{The slot accuracy of our model and BERT-DST on the modified test set. The middle column presents all unenumerated slots in Sim-M and Sim-R.}
	\label{generalization_study}
\end{figure}

To investigate the limitations of our model in terms of generalization ability, we manually replace all unenumerated slot values in the test set of Sim-M and Sim-R with unknown slot values (manually written but meaningful). The slot accuracy of our model and BERT-DST on the modified test set is shown in Figure \ref{generalization_study}. It can be observed that the accuracy of most slots of our model is above 90\%, with the lowest remaining at 87.9\%. This indicates that our model is well equipped to handle unknown slot values and its generalization is guaranteed. 

Compared with BERT-DST, our model has great advantages in handling unknown slot values with longer lengths, i.e., $movie$, $theatre\_name$, and $restaurant\_name$, which on average contain 2.8, 3.4 and 1.6 words in the test set. Since these slot values are more likely to appear in the form of unknown and complex representation in practice, the results of our model demonstrate that our model also has a fantastic potential in practical application.

\section{Conclusion}
We propose STN4DST, a scalable dialogue state tracking approach based on slot tagging navigation, which uses slot tagging to accurately locate candidate slot values in dialogue content, and then uses the single-step pointer to quickly extract the slot values. STN4DST achieves state-of-the-art joint goal accuracy on both Sim-M and Sim-R while maintains comparable results on WOZ2.0 and MWOZ2.2. When dealing with more complex unknown slot values, STN4DST presents better generalization and scalability than the widely used span extraction, showing greater research potential and application prospect.

\bibliographystyle{acl_natbib}
\bibliography{references}


\end{document}